\PassOptionsToPackage{unicode}{hyperref}
\PassOptionsToPackage{hyphens}{url}
\documentclass[]{article}
\usepackage{amsmath,amssymb}
\usepackage[letterpaper]{geometry}
\usepackage{iftex}
\ifPDFTeX
  \usepackage[T1]{fontenc}
  \usepackage[utf8]{inputenc}
  \usepackage{textcomp} 
\else 
  \usepackage{unicode-math} 
  \defaultfontfeatures{Scale=MatchLowercase}
  \defaultfontfeatures[\rmfamily]{Ligatures=TeX,Scale=1}
\fi
\usepackage{lmodern}
\ifPDFTeX\else
\fi
\IfFileExists{upquote.sty}{\usepackage{upquote}}{}
\IfFileExists{microtype.sty}{
  \usepackage[]{microtype}
  \UseMicrotypeSet[protrusion]{basicmath} 
}{}
\makeatletter
\@ifundefined{KOMAClassName}{
  \IfFileExists{parskip.sty}{%
    \usepackage{parskip}
  }{
    \setlength{\parindent}{0pt}
    \setlength{\parskip}{6pt plus 2pt minus 1pt}}
}{
  \KOMAoptions{parskip=half}}
\makeatother
\usepackage{xcolor}
\usepackage{longtable,booktabs,array}
\usepackage{calc} 
\usepackage{etoolbox}
\makeatletter
\patchcmd\longtable{\par}{\if@noskipsec\mbox{}\fi\par}{}{}
\makeatother
\IfFileExists{footnotehyper.sty}{\usepackage{footnotehyper}}{\usepackage{footnote}}
\makesavenoteenv{longtable}
\usepackage{graphicx}
\makeatletter
\def\maxwidth{\ifdim\Gin@nat@width>\linewidth\linewidth\else\Gin@nat@width\fi}
\def\maxheight{\ifdim\Gin@nat@height>\textheight\textheight\else\Gin@nat@height\fi}
\makeatother
\setkeys{Gin}{width=\maxwidth,height=\maxheight,keepaspectratio}
\makeatletter
\def\fps@figure{htbp}
\makeatother
\ifLuaTeX
  \usepackage{selnolig}  
\fi
\IfFileExists{bookmark.sty}{\usepackage{bookmark}}{\usepackage{hyperref}}
\IfFileExists{xurl.sty}{\usepackage{xurl}}{} 
\hypersetup{
  hidelinks,
  pdfcreator={LaTeX via pandoc}}

\author{}
\date{}

\begin{document}
\newgeometry{bottom=3cm}
\hypertarget{improving-accuracy-of-gpt-34-results-on-biomedical-data-using-a-retrieval-augmented-language-model}{%
\subsection{\texorpdfstring{\textbf{Improving accuracy of GPT-3/4
results on biomedical data using a retrieval-augmented language
model}}{Improving accuracy of GPT-3/4 results on biomedical data using a retrieval-augmented language model}}\label{improving-accuracy-of-gpt-34-results-on-biomedical-data-using-a-retrieval-augmented-language-model}}

David Soong*\textsuperscript{1}, Sriram Sridhar*\textsuperscript{1}, Han
Si\textsuperscript{1}, Jan-Samuel Wagner\textsuperscript{1}, Ana
Caroline Costa Sá\textsuperscript{1}, Christina Y Yu\textsuperscript{1},
Kubra Karagoz\textsuperscript{1}, Meijian Guan\textsuperscript{1},
Hisham Hamadeh\textsuperscript{1}, Brandon W Higgs\textsuperscript{1}

\textsuperscript{1}Data Sciences, Genmab, Princeton, NJ

*Equal contribution

\hypertarget{abstract}{%
\subsection{\texorpdfstring{\textbf{Abstract}}{Abstract}}\label{abstract}}

Large language models (LLMs) have made significant
advancements in natural language processing (NLP). Broad corpora capture
diverse patterns but can introduce irrelevance, while focused corpora
enhance reliability by reducing misleading information. Training LLMs on
focused corpora poses computational challenges. An alternative approach
is to use a retrieval-augmentation (RetA) method tested in a specific
domain.

To evaluate LLM performance, OpenAI\textquotesingle s GPT-3.5,
GPT-4, Bing\textquotesingle s Prometheus, and a custom RetA model were
compared using 19 questions on diffuse large B-cell lymphoma (DLBCL)
disease. Eight independent reviewers assessed responses based on
accuracy, relevance, and readability (rated
1-3).

The RetA model performed best in accuracy (12/19 3-point
scores, total=47) and relevance (13/19, 50), followed by GPT-4 (8/19,
43; 11/19, 49). GPT-4 received the highest readability scores (17/19,
55), followed by GPT-3.5 (15/19, 53) and the RetA model (11/19, 47).
Prometheus underperformed in accuracy (34), relevance (32), and
readability
(38).

Both GPT-3.5 and GPT-4 had more hallucinations in all 19
responses compared to the RetA model and Prometheus. Hallucinations were
mostly associated with non-existent references or fabricated efficacy
data.

These findings suggest that RetA models, supplemented with
domain-specific corpora, may outperform general-purpose LLMs in accuracy
and relevance within specific domains. However, this evaluation was
limited to specific questions and metrics and may not capture challenges
in semantic search and other NLP tasks. Further research will explore
different LLM architectures, RetA methodologies, and evaluation methods
to assess strengths and limitations more
comprehensively.

\hypertarget{section}{%
\subsection{}\label{section}}

\hypertarget{introduction}{%
\subsection{\texorpdfstring{\textbf{Introduction}}{Introduction}}\label{introduction}}

The development of large language models (LLMs), such as bidirectional
encoder representations from transformer (BERT) and generative
pre-trained transformer (GPT), has revolutionized the field of natural
language processing {[}1{]}, {[}2{]}, {[}3{]} {[}4{]}. Applications of
these LLMs have ranged from sentiment analysis and machine translation
to code generation and question answering in several domains {[}5-10{]}
-- all demonstrating remarkable performance. However, despite their
impressive execution and widespread use, LLMs do not know the
information they were not trained on, and often lack domain-specific
knowledge and vocabulary. They can also perpetuate biases based on
skewed content in the training data, and need to be further refined
through reinforcement learning and alignment approaches to understand
user intentions while making them more truthful and less toxic {[}11,
12{]}. Furthermore, concerns have been raised about the potential for
LLMs to generate hallucinated or misleading information, which can have
severe implications in scientific research and led to the critical
determinants of distinguishing fact from fiction leading to
discontinuation of, as was the case for Meta's Galactica {[}13, 14{]}.

Popular LLMs with billions of parameters such as GPT-3 {[}4{]} , PaLM
{[}15{]}, OPT {[}16{]}, and LLaMA {[}17{]} are typically trained on vast
amounts of information collected from the Internet (e.g. the Common
Crawl dataset {[}18{]}) and capture a diverse range of language patterns
and knowledge. Word and sentence embeddings are high-dimensional
numerical representations of concepts scaled by the size of the corpus
and complexity of language usage {[}19{]} {[}20, 21{]}. This can produce
a higher level of generality and flexibility in the
model\textquotesingle s ability to yield natural language, making it
more robust and adaptable to a range of applications. Similarly, a broad
corpus can capture the diversity of language usage across different
domains and genres. For example, a model trained on a broad corpus could
potentially generate natural language in scientific literature, social
media, or news articles, with equal ease {[}3{]}.

Nonetheless, a wide-ranging corpus can inadvertently incorporate a
significant amount of noise or irrelevant data, resulting in a reduced
signal-to-noise ratio {[}22{]}. This may adversely affect the generated
text\textquotesingle s quality, leading to decreased coherence, meaning,
or accuracy. Additionally, biases and inaccuracies may arise in the
model\textquotesingle s comprehension of natural language. A corpus that
predominantly features one type of language or cultural context may
display bias towards that specific domain or culture {[}23{]}. Although
a corpus may strive to encompass a diverse range of domains, the sheer
vastness of the domain space makes it currently unfeasible to include
all relevant domains. Moreover, as more domains are incorporated, there
is a risk that LLMs trained on such a comprehensive corpus may struggle
to differentiate language from various domains, particularly when faced
with prompts that lack sufficient context.

One approach to address these limitations is to retrain or finetune an
LLM with a focused corpus tailored to a specific domain or application
{[}22{]} {[}24{]} thereby reducing the risk of generating irrelevant or
misleading information and enhancing the reliability and precision of
the LLM\textquotesingle s outputs in specialized contexts. Numerous
publications have highlighted the efficacy of domain specific LLMs in
their respective fields. For example, BioBERT {[}25{]} targets
biomedical text mining tasks, SciBERT {[}26{]} and PubMedBERT {[}27{]}
address scientific literature, and Legal-BERT {[}28{]} specializes in
legal text processing. These approaches minimize noise and irrelevant
information in the text, potentially reducing hallucinations.

However, retraining LLMs to encompass new documents might be impractical
due to the cumulative computational costs and data scientist resources
required per update. The LLM architecture might also need to be updated
to incorporate more parameters to memorize more facts {[}29{]}. As LLMs
have demonstrated extraordinary abilities to learn in-context
information purely from its prompt {[}4{]}, RetA approaches have proven
promising {[}24{]} {[}30{]}. These models first retrieve relevant
context from domain-specific corpora based on a user query using lexical
search (e.g. BM25 {[}31{]}) or a pretrained/fine-tuned semantic
retriever (e.g. Spider{[}32{]}, OpenAI embeddings {[}33{]}), and then
seed a pre-trained LLM with such context to provide grounded answers
while avoiding the prohibitive time and cost of retraining an LLM.

In this study, several LLMs were evaluated to investigate if a
retrieval-augmentation approach on a focused corpus could improve the
accuracy of LLMs applications in biomedical Q\&A. Three scoring metrics
were utilized to objectively compare outputs between models using a set
of evaluation-based questions focused on disease characterization,
genetic subtypes, treatment options, and clinical outcomes in diffuse
large B-cell lymphoma DLBCL. These observations provide insights into
the pros and cons of each LLM and suggest potential areas for
improvement to meet utility requirements for rigorous drug development
and scientific research.

\hypertarget{methods}{%
\subsection{\texorpdfstring{\textbf{Methods}}{Methods}}\label{methods}}

\emph{\textbf{Evaluation framework}}

The performance of generically trained LLMs was tested versus a RetA LLM
in question answering (Q\&A) tasks related to disease biology and drug
development. A set of 19 questions focused on mechanisms and treatments
associated with DLBCL were provided to evaluate LLM performance. The
questions covered a broad range of topics related to DLBCL disease
biology including clinical and molecular subtypes, genetic subsets and
relevant biomarkers, clinical management, and standards of care and
other available therapies. Questions were designed to look for both
qualitative and quantitative answers (e.g. overall response rate and
prevalence of genomic alterations). Each question was provided to four
different LLMs: Open AI's general ChatGPT-3.5 {[}34{]}, OpenAI's general
GPT-4 {[}34{]}, Bing's Prometheus model (referred to in this manuscript
as Bing chat, based on GPT-4 {[}35{]}), and a RetA LLM (based on GPT-3)
using a custom set of full-text publications associated with DLBCL
(\textbf{Table 1}). The questions intentionally varied in detail to
assess the ability of each LLM to infer the expected result. For
example, question \#15 provided a concise query for DLBCL diagnosis and
prognosis, while question \#3 asked specific treatments for a target in
the disease with accompanying references to support the answer.

The two general GPT-based LLMs from OpenAI were only trained on content
up to September 2021 (OpenAI GPT-4 Technical Report {[}36{]}), as
opposed to Bing's Prometheus and the RetA models. Release versions of
GPT-4 and GPT-3.5 used to answer the questions were from 3/23/23 to
4/28/23 (updates were released on a weekly or bi-weekly basis and were
documented).

\emph{\textbf{RetA model and dataset}}

Scientific papers were downloaded from PubMed Central (PMC {[}37{]})
using the Entrez E-utilities {[}38{]}). Each of the following queries
was used to retrieve up to 500 articles: `diffuse large b-cell
lymphoma', `follicular lymphoma', `epcoritamab', `glofitamab', `minimal
residual disease', `ctDNA'. By default, Entrez returns articles sorted
by PMC identifier. The queries used were meant to generate a corpus
specific to DLBCL, related biomarkers, standards of care, and
therapeutic options, not to specifically answer the questions used in
this evaluation. This created a unique dataset of 1,868 full-text
articles. The documents were first pre-processed to exclude potentially
unstructured or noisy text (e.g. figures, tables, references, author
disclosure) and split into segments of 4,000 tokens. Embeddings were
then calculated using the OpenAI model text-embedding-ada-002 and stored
in a local database. When the user entered a question, the query was
transformed into an embedding vector and compared to the database of
embeddings using cosine similarity. The top \emph{k} document segments
by similarity were retrieved and formed the knowledge context for the
user query. The synthesis of the answer to the query was achieved in two
stages: in stage one, text-davinci-003 was used to answer the query
using each of the \emph{k} context segments with prompt instructions to
minimize inclusion of non-factual information from the LLM. This
generated \emph{k} answers which were combined into a final response in
the second stage using a call to text-davinci-003 with a summarization
prompt (\textbf{Figure 1, Tables 2a,b}).

\emph{\textbf{Evaluation metrics}}

Answers were scored for each question on a three-point scale (1-3, with
3 being highest) based on three metrics: accuracy, relevance, and
readability by eight independent reviewers (\textbf{Table 3}), with each
reviewer scoring a subset of questions. Answers to all questions were
searchable. Accuracy and relevance assessments focused on factual
correctness of answers, correctness of references or links to
references, or general pieces of knowledge included or not included in
an answer. The 3-point scale used for each evaluation category also
allowed for some granularity in scoring answers. For example, an answer
might be given a score of ``2'' if the result was factually correct but
links to supporting references were broken or incorrect. An answer which
does not directly address the question being asked or contains factually
incorrect information (i.e. hallucinations) might garner a score of
``1'' for accuracy. As both the language model and oncology therapeutics
fields are constantly evolving, there is some recency bias associated
with answers to questions and the data which LLMs are trained on. This
was in part accounted for through the types of questions chosen and the
scale used to assess responses. An emphasis of the evaluation was to
specifically look for factually incorrect answers, as opposed to
incomplete answers which may be a result of recency bias. Reviewers were
all Ph.D. level scientists with an average of 8 years of biopharma
industry experience and 11 years of post-doctoral work experience. All
scores were then assessed by one reviewer from the group to adjust for
reviewer biases. The prompts were stratified into three high level
categories based on relevance to drug information, disease biology, and
clinical information. Prompts were also grouped based on being general
(i.e. high level) or specific (i.e. asking for details) questions to
better attribute subfield performance within DLBCL in comparisons
between LLMs.

\hypertarget{results}{%
\subsection{\texorpdfstring{\textbf{Results}}{Results}}\label{results}}

Overall, the performance of the LLMs varied widely across the different
questions and metrics. In terms of accuracy, the RetA model of GPT-3 on
DLBCL publications outperformed the other LLMs with the highest
(3-point) scoring answers on 12/19 questions. GPT-4 was the next best
performer with 3-point scores on 8/19 questions. Bing's Prometheus had
7/19 3-point scores for accuracy while GPT-3.5 had the fewest high
scoring answers (4/19 3-point scores) (\textbf{Figures 2, 3}). The
summated scores for accuracy showed that the RetA model scored slightly
higher than GPT-4 in the categories of drug and clinical information
(\textbf{Figure 4}). Bing's Prometheus model did not perform well in
accuracy compared to all other models with low (1-point) scores on 10/19
questions \textbf{(Figures 2, 3)}. This was primarily due to
misrepresentation of references in its answers. Conversely, GPT-4 and
the RetA model had the fewest low scoring answers for accuracy (1/19 and
3/19 respectively) across prompts \textbf{(Figures 2, 3)}.

Interestingly, Bing's Prometheus model was the only one to not score a
value of 1 in accuracy for question \#6 (``What is the overall response
rate of DLBCL patients treated with glofitamab?''). Numerical overall
and complete response rates (ORR and CRR, respectively) reported by
GPT-3.5 (ORR=65.1\%, CRR=35.1\%) and GPT-4 (ORR=62.7\%, CRR=39.2\%) were
not consistent with their references cited and had either fabricated or
provided incorrect references. Bing's Prometheus model scored a value of
2 because there was a mixture of accurate and inaccurate answers to the
question, i.e., this model accurately captured the ORR value of
glofitamab treatment (52\%) in Dickinson et al, NEJM reference {[}39{]},
but also incorrectly used the median duration of objective response
rather than median duration of CR. The RetA model result was not
accurate in answering this question because the official glofitamab
trial efficacy paper {[}39{]} was not available on PubMed Central
(\url{https://www.ncbi.nlm.nih.gov/pmc/}) and therefore not included in
the corpus.

In terms of relevance, the RetA model performed slightly better than
GPT-4 and GPT-3.5. The RetA model scored high (3-point) on 13/19
questions, compared to 11/19 and 10/19 in GPT-4 and GPT-3.5
respectively. Bing's Prometheus model performed worst in this category
with scores of 1 in 8/19 questions \textbf{(Figures 2, 3)}. The other
three LLMs had few-to-no low scoring answers to prompts with respect to
relevance. The irrelevant answers (i.e. low scoring questions) across
all LLMs were primarily due to references to other diseases or
treatment. For example, in question \#14 (``Have checkpoint inhibitor
treatments in monotherapy or combination therapy settings shown efficacy
in DLBCL patients? Provide references.''), the GPT-4 model cited three
references, one of which was in Hodgkin's lymphoma (DLBCL is a
non-Hodgkin's lymphoma) and another that discussed CAR-T, which is not a
checkpoint inhibiting drug agent, though the model associated this
treatment modality with immunotherapies and extended relevance to CAR-T
therapies. GPT-3.5 also cited a reference evaluating a checkpoint
inhibitor treatment in Hodgkin's lymphoma.

Finally, for readability, GPT-4 scored the highest with 17/19 scores of
3, followed by GPT-3.5 with 15/19, and the RetA model with 11/19
\textbf{(Figures 2, 3)}. The summated scores demonstrated parity between
GPT-4 and GPT-3.5 across all categories (\textbf{Figure 4}). Readability
was particularly low scoring in the clinical category of questions for
the RetA. Bing's Prometheus model once again scored last in this
category (7/19 3-point scores), primarily due to concise, yet vague
answers, often with little detail. For example, for question \#7 (``What
is a treatment to use in DLBCL patients who have progressed on
CAR-T?''), Bing's Prometheus model simply reported references without
summarization, including one study where multiple drugs were approved,
and referenced only those of approved agents, ignoring studies
evaluating investigational drug agents.

Across the 19 questions, both GPT-3.5 and GPT-4 LLMs generated a
considerably higher number of hallucinations in their responses (31 from
13 questions and 19 from 8 questions, respectively) compared to the RetA
model and Bing' Prometheus model (3 from 3 questions and 2 from 1
question, respectively). These were primarily associated with
fabrication of both references and clinical results. Although LLMs are
known to be behind in mathematical capabilities {[}40{]}, the inaccuracy
of numerical results appeared to be due to hallucinations or context
understanding rather than limitations in mathematical reasoning.

These results suggest that the performance of LLMs can vary widely
depending on the specific task and domain, though the RetA model
enhanced with domain-specific data may outperform more general-purpose
LLMs in accuracy and relevance. However, it should be noted that this
evaluation was limited to a specific set of questions and metrics, and
further research is needed to fully understand the strengths and
limitations of different LLMs for semantic search and other natural
language processing tasks.

\hypertarget{discussion}{%
\subsection{\texorpdfstring{\textbf{Discussion}}{Discussion}}\label{discussion}}

The advantages and drawbacks of using LLMs trained on broad corpora
versus a RetA approach ultimately depend on the specific use case and
desired outcomes. In biomedical and healthcare research, it is paramount
to have accurate, relevant, and unbiased information supported by
published literature. In this study, quantifying the accuracy and
utility of LLMs was conducted for answering qualitative and quantitative
biomedical questions related to the treatment and prognosis of patients
with DLBCL. Results here demonstrated that the RetA LLM performed better
on biomedical-specific tasks than the other LLMs evaluated, specifically
with respect to accuracy of results. This suggests that RetA LLMs can
provide more accurate and reliable information for specific fields,
reducing the likelihood of generating irrelevant or misleading outputs,
while maintaining the flexibility and adaptability of a general LLM.

One major advantage of the RetA model is the easy integration of new
domain knowledge that the base LLM was not trained on. When a new
document is added to the corpus, the model only needs to calculate the
embeddings to facilitate retrieval during future queries. On the other
hand, fine-tuning or retraining an LLM on a new corpus takes both time
and resources, and may not always be possible depending on the choice of
LLM -- as of the publication of this study, OpenAI has not offered an
option to fine-tune their ChatGPT models; Meta's LLaMA model is also not
available for commercial applications {[}17, 41{]}.

However, since the RetA model needs to prompt a pre-trained LLM into
performing specific tasks such as summarizing across relevant documents
and extracting information without using prior knowledge, the model
typically uses a large amount of tokens as input and multiple iterations
of base LLM inference (i.e. text-completion API) calls, which can
increase the compute cost in its application. The dependence on a
certain LLM (e.g. OpenAI GPT-3) also implies that the desired prompt
behavior needs to be closely monitored when the LLM backend is updated
with new training data, or when the user switches to a different base
LLM (e.g. GPT-4, Dolly 2 {[}42{]}, Open Assistant {[}43{]}, or RedPajama
{[}44{]}).

Furthermore, its performance is also bound by the limitations of the
base LLM's vocabulary (tokenizer) and internal representation of
concepts (embedding). For example, question \#13 asked about minimal
residual disease (MRD) in DLBCL, but the document retriever returned
articles about MRD in multiple myeloma and chronic lymphocytic leukemia
- two distinct hematological malignancies from DLBCL. The RetA model
relies on GPT-3 as the summarization engine which failed to distinguish
between the different disease types, leading to an incorrect answer.
These issues may be ameliorated by utilizing more sophisticated document
retrieval methods. For biomedical literature, domain specific models
such as BioBERT and PubMedBERT can be used for tokenization and
embedding calculation; additional metadata filters can also be used to
improve relevance of retrieved documents. As an example, when the
retrieval method was modified in the RetA model to directly search for
supporting articles on PubMed by significance, the model provided
informative and relevant answers detailing the measurement of disease
clones with V(D)J sequences, as well as the association with clinical
outcomes.

Overall, general LLMs provide highly readable and coherent text in
various subjects. Furthermore, the performance of the RetA model
demonstrated the utility of using LLMs as a backend in performing
various reasoning tasks through specifically crafted prompts. Indeed,
prompt engineering has been an active area of research that continues to
expand the capability of pre-trained LLMs through methods such as:
zero-shot {[}45{]}, few-shot {[}4{]}, chain of thought {[}46{]},
self-ask {[}47{]}, and ReAct {[}48{]} reasoning. These reasoning
properties allow LLMs to be used as programmable agents to orchestrate
and perform tasks across different modalities or domains (e.g.
ToolFormer {[}49{]}, Visual ChatGPT {[}50{]}, Langchain {[}51{]}, GPT
plugins {[}52{]}).

Though findings here are informative, this study had several limitations
that need to be considered. First, the assessment included only 19
questions, which accounted for various clinical, therapy, and biological
content, which was an attempt to address pertinent context in biomedical
research, though certainly not exhaustive. Second, the focus was on a
single disease (DLBCL), which may not be generalizable to other diseases
or domains. Third, the scoring metrics selected included accuracy,
readability, and relevance, which might not have captured other
important aspects of the text such as strength, completeness, and
consistency. The scoring was performed across the entire answer as
opposed to by sentence or phrase within an answer. While scoring
questions in this manner can be subjective, we adjusted for this by
using multiple reviewers and having an additional overarching review to
calibrate scores across questions. There was also a range in experience
among reviewers to account for any bias associated with experience.
Questions were also specific enough such that available literature could
be used to assess accuracy of answers. A point of emphasis for the
evaluation of responses was to look for factually incorrect answers
(hallucinations), which were more likely to garner the lowest score, as
opposed to answers which were factually correct but not exhaustive.
Last, the RetA model included an arbitrary number of full-text articles
(1,868), prioritized by PMC identifier, which might not have represented
the most relevant or comprehensive set of articles for the disease. It
is possible that an optima of accuracy, relevance, and readability can
be achieved with an RetA model by increasing the size and breadth of the
corpus, and future work will be needed to test this hypothesis. Despite
these limitations, this study provides valuable insights into the
performance of LLMs on different types of corpora and highlights the
importance of domain-specific knowledge in achieving higher accuracy and
relevance.

With the rapid advancement and development of foundation models across
text, image, video and other data modalities, adaptation of AI in a
fair, accurate, and reliable fashion can make an immediate impact on
healthcare and drug development. In this study, focus was on evaluation
of pre-trained and RetA LLMs for biomedical Q\&A in the field of
clinical drug development. Future research could explore methods
incorporating biomedical ontology, knowledge graphs, as well as other
agent-based approaches to further enhance the performance of LLMs
{[}51{]}, {[}52{]}. As open-source initiatives democratize AI research
{[}53, 54{]}{[}42{]} {[}43{]} {[}44{]} and new emerging methodologies
{[}55-58{]} begin to offer possibilities to build custom LLMs with
reduced compute resources and time requirements, further integrating
with multi-modal approaches that leverage across molecular (e.g.
mutations and gene expression), imaging (pathology and radiology),
electronic health records, and wearable sensor data will provide a
deeper understanding of disease biology and accelerate drug development
in a fair and socially responsible way {[}59{]} {[}60{]}.

\hypertarget{acknowledgements}{%
\subsection{\texorpdfstring{\textbf{Acknowledgements}}{Acknowledgements}}\label{acknowledgements}}

The authors thank Bryan Ho and Swathi Vangala for their contributions to
infrastructure.

\hypertarget{figures-and-tables}{%
\subsection{\texorpdfstring{\textbf{Figures and
Tables}}{Figures and Tables}}\label{figures-and-tables}}

\includegraphics{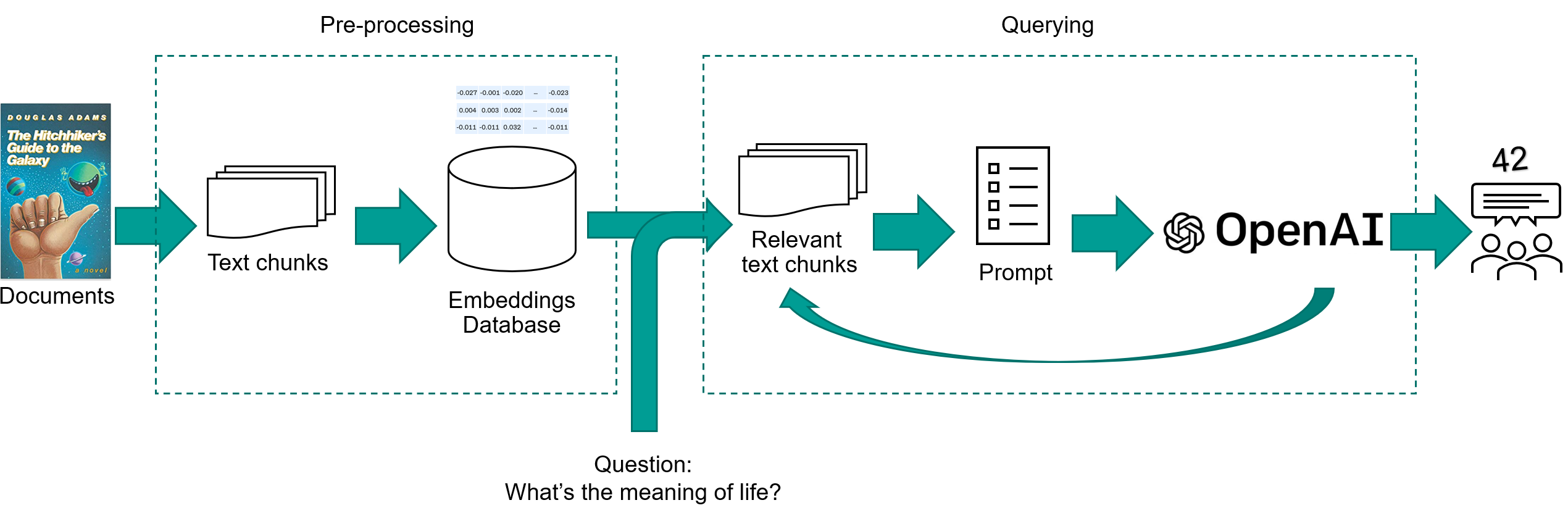}

\textbf{Figure
1}. Components and workflow of a RetA LLM. The pre-processing stage
splits documents into smaller chunks, creates embeddings and stores them
in a database. At the querying stage, a document retriever finds the
most relevant documents in the embeddings database, iteratively seeds
the base LLM with context to generate a response.

\textbf{Table 1}. Questions used for LLM evaluation classified into
group and scope categories.

\begin{longtable}[]{@{}
  >{\raggedright\arraybackslash}p{(\columnwidth - 6\tabcolsep) * \real{0.1651}}
  >{\raggedright\arraybackslash}p{(\columnwidth - 6\tabcolsep) * \real{0.5426}}
  >{\raggedright\arraybackslash}p{(\columnwidth - 6\tabcolsep) * \real{0.1909}}
  >{\raggedright\arraybackslash}p{(\columnwidth - 6\tabcolsep) * \real{0.1013}}@{}}
\toprule\noalign{}
\begin{minipage}[b]{\linewidth}\raggedright
Question \#
\end{minipage} & \begin{minipage}[b]{\linewidth}\raggedright
Question
\end{minipage} & \begin{minipage}[b]{\linewidth}\raggedright
Group
\end{minipage} & \begin{minipage}[b]{\linewidth}\raggedright
Scope
\end{minipage} \\
\midrule\noalign{}
\endhead
\bottomrule\noalign{}
\endlastfoot
1 & What is epcoritamab? Please provide sources for your answer. & Drug
information & General \\
2 & What are the subtypes of DLBCL? Please provide sources for your
answer. & Disease biology & General \\
3 & What are the antibody therapies targeting CD20 for treatment of
DLBCL? Please provide sources for your answer. & Drug information &
General \\
4 & What is the standard of care for treatment of DLBCL? & Clinical
information & Specific \\
5 & What are the approved drugs for treatment of DLBCL? & Clinical
information & Specific \\
6 & What is the overall response rate of DLBCL patients treated with
glofitamab? & Clinical information & Specific \\
7 & What is a treatment to use in DLBCL patients who have progressed on
CAR-T? & Drug information & General \\
8 & What are common treatments used in patients who have relapsed or
were refractory to standard of care treatments in DLBCL? & Drug
information & General \\
9 & Do any DLBCL patient subtypes respond more favorably to chemotherapy
or CAR-T treatments? & Clinical information & Specific \\
10 & What are the most common adverse events observed in DLBCL patients
treated with R-CHOP? & Clinical information & Specific \\
11 & What biomarkers in DLBCL have been reported to correlate with
either response or progression following treatment with R-CHOP? &
Clinical information & Specific \\
12 & What treatment combinations have been shown to be effective in
DLBCL patients who have progressed on CAR-T treatment? Please provide
sources for your answer. & Clinical information & Specific \\
13 & How can minimal residual disease (MRD) be used to understand
clinical outcomes in DLBCL patients? Please provide sources for your
answer. & Disease biology & General \\
14 & Have checkpoint inhibitor treatments in monotherapy or combination
therapy settings shown efficacy in DLBCL patients? Provide references. &
Drug information & Specific \\
15 & DLBCL diagnosis and prognosis. & Clinical information & General \\
16 & Landscape of DLBCL treatment as SOC. Please provide sources for
your answer. & Clinical information & Specific \\
17 & Emerging novel treatment options for DLBCL patients. & Drug
information & General \\
18 & what is the importance of TP53 in DLBCL? & Disease biology &
General \\
19 & What is the prevalence of double hit mutations in lymphoma? &
Disease biology & Specific \\
\end{longtable}

\textbf{Table 2a}. Prompts for GPT3 in the RetA workflow.

\begin{longtable}[]{@{}
  >{\raggedright\arraybackslash}p{(\columnwidth - 2\tabcolsep) * \real{0.0940}}
  >{\raggedright\arraybackslash}p{(\columnwidth - 2\tabcolsep) * \real{0.9060}}@{}}
\toprule\noalign{}
\begin{minipage}[b]{\linewidth}\raggedright
\textbf{Stage}
\end{minipage} & \begin{minipage}[b]{\linewidth}\raggedright
\textbf{Prompt}
\end{minipage} \\
\midrule\noalign{}
\endhead
\bottomrule\noalign{}
\endlastfoot
Stage one & Instruction: You are a truthful AI assistant. You answer
questions only based on provided context below. If the context is not
relevant to the question, say you do not know the answer. No need to
explain why.

Context: \{segment of article\}

Question: \{user query\}

Answer: \\
Stage two & Please combine the following paper\textquotesingle s
summaries. Only use the context below and not incorporate any prior
knowledge.

Paper \#1: \{answer 1 based on segment 1\}

Paper \#2: \{answer 2 based on segment 2\} \\
\end{longtable}

\textbf{Table 2b}. Workflow and LLM descriptions used in this study.

\begin{longtable}[]{@{}
  >{\raggedright\arraybackslash}p{(\columnwidth - 4\tabcolsep) * \real{0.3314}}
  >{\raggedright\arraybackslash}p{(\columnwidth - 4\tabcolsep) * \real{0.2344}}
  >{\raggedright\arraybackslash}p{(\columnwidth - 4\tabcolsep) * \real{0.4341}}@{}}
\toprule\noalign{}
\begin{minipage}[b]{\linewidth}\raggedright
\textbf{Workflow}
\end{minipage} & \begin{minipage}[b]{\linewidth}\raggedright
\textbf{Evaluation}
\end{minipage} & \begin{minipage}[b]{\linewidth}\raggedright
\textbf{Base LLM}
\end{minipage} \\
\midrule\noalign{}
\endhead
\bottomrule\noalign{}
\endlastfoot
RetA LLM & Python workflow & text-davinci-003 \\
chatGPT3.5 & OpenAI web & gpt-3.5-turbo \\
chatGPT4 & OpenAI web & gpt-4 \\
BingChat & Microsoft web & Custom GPT4 \\
\end{longtable}

\textbf{Table 3}. Answer scoring metric descriptions for LLM comparison.

\begin{longtable}[]{@{}
  >{\raggedright\arraybackslash}p{(\columnwidth - 6\tabcolsep) * \real{0.1917}}
  >{\raggedright\arraybackslash}p{(\columnwidth - 6\tabcolsep) * \real{0.2694}}
  >{\raggedright\arraybackslash}p{(\columnwidth - 6\tabcolsep) * \real{0.2694}}
  >{\raggedright\arraybackslash}p{(\columnwidth - 6\tabcolsep) * \real{0.2694}}@{}}
\toprule\noalign{}
\begin{minipage}[b]{\linewidth}\raggedright
\end{minipage} &
\multicolumn{3}{>{\raggedright\arraybackslash}p{(\columnwidth - 6\tabcolsep) * \real{0.8083} + 4\tabcolsep}@{}}{%
\begin{minipage}[b]{\linewidth}\raggedright
\textbf{Score}
\end{minipage}} \\
\midrule\noalign{}
\endhead
\bottomrule\noalign{}
\endlastfoot
\textbf{Metrics} & \textbf{1} & \textbf{2} & \textbf{3} \\
Accuracy & Mostly inaccurate or misleading content & A mix of accurate
and inaccurate content & Factually accurate and reliable content \\
Relevance & Mostly irrelevant content & Partially relevant content &
Highly relevant and on-point content \\
Readability & Difficult to read, unclear or convoluted language &
Moderately readable, with some unclear passages & Easy to read, clear,
and concise language \\
\end{longtable}

\includegraphics{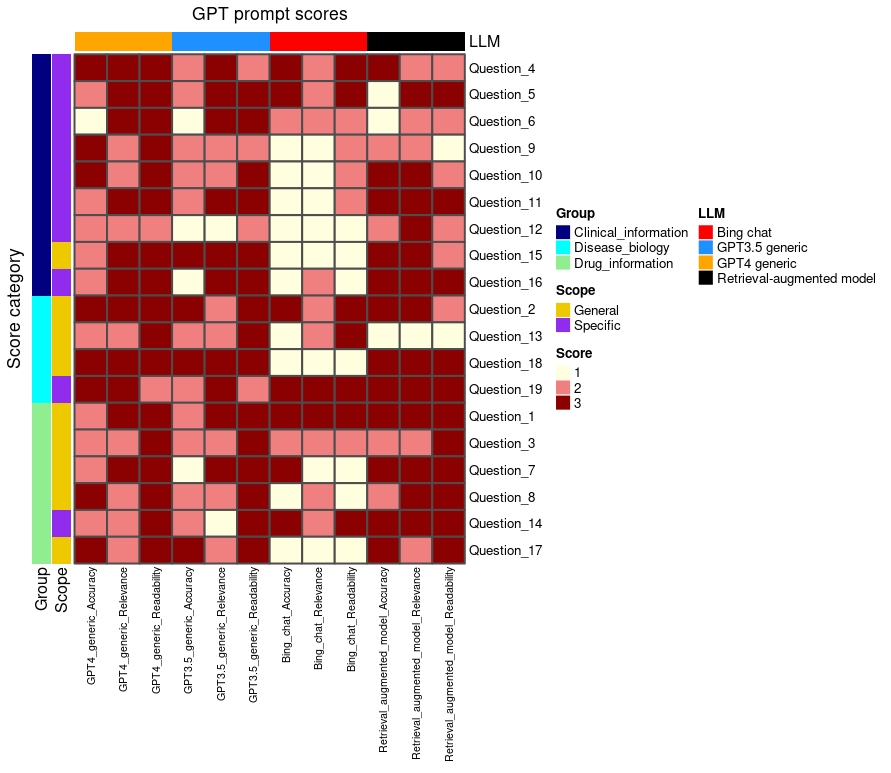}

\textbf{Figure 2}. Scores for each LLM within 3 metrics (accuracy,
relevance, readability) on a three-point scale. Questions are ordered by
question category (clinical, drug-related, disease-related). Question
scope (general or specific) is also annotated.

\includegraphics{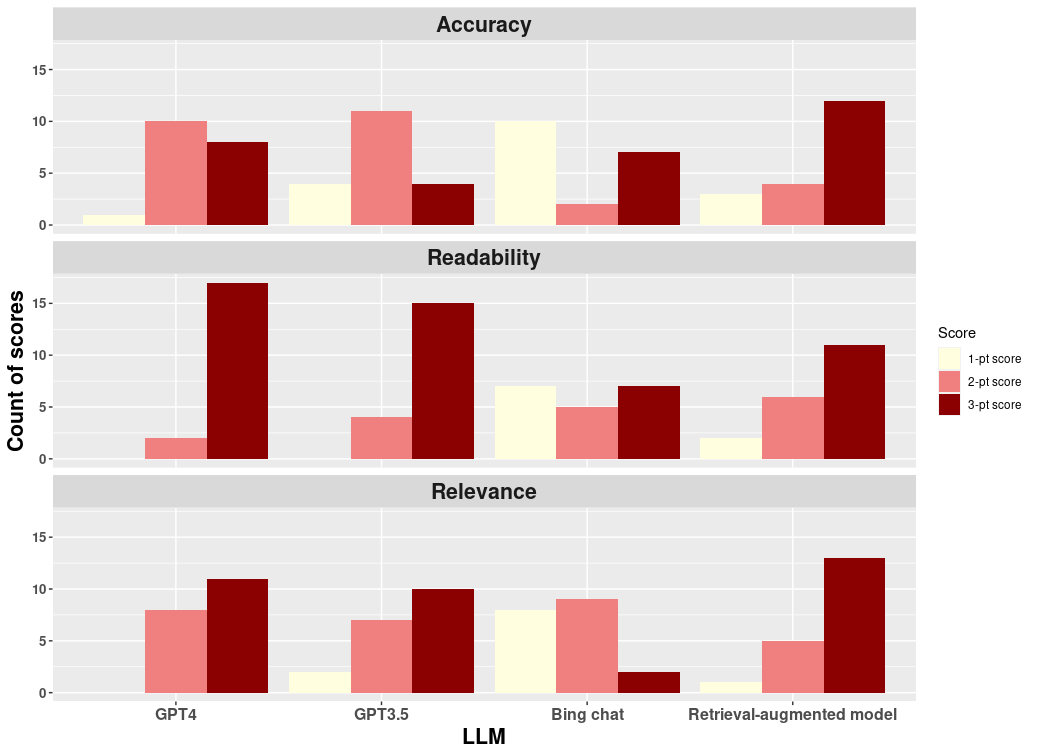}

\textbf{Figure 3}. Count of scores (3-point, 2-point, and 1-point)
across the 19 questions for each LLM in each score category (Accuracy,
Readability, Relevance).

\includegraphics{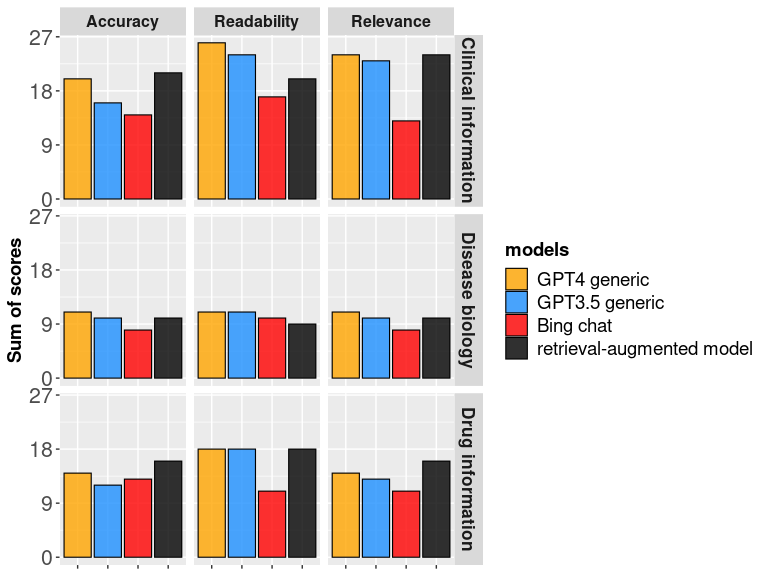}

\textbf{Figure 4}. Summarized scores for each LLM within 3 metrics
(accuracy, relevance, readability) and question categories (clinical
information, disease biology, and drug information).

\hypertarget{references}{%
\subsection{\texorpdfstring{\textbf{References}}{References}}\label{references}}

1. Vaswani, A., et al. \emph{Attention Is All You Need}. 2017.
arXiv:1706.03762 DOI: 10.48550/arXiv.1706.03762.

2. Devlin, J., et al. \emph{BERT: Pre-training of Deep Bidirectional
Transformers for Language Understanding}. 2018. arXiv:1810.04805 DOI:
10.48550/arXiv.1810.04805.

3. Radford, A., et al. \emph{Language Models are Unsupervised Multitask
Learners}. 2019.

4. Brown, T.B., et al. \emph{Language Models are Few-Shot Learners}.
2020. arXiv:2005.14165 DOI: 10.48550/arXiv.2005.14165.

5. Chen, M., et al. \emph{Evaluating Large Language Models Trained on
Code}. 2021. arXiv:2107.03374 DOI: 10.48550/arXiv.2107.03374.

6. Gozalo-Brizuela, R. and E.C. Garrido-Merchan \emph{ChatGPT is not all
you need. A State of the Art Review of large Generative AI models}.
2023. arXiv:2301.04655 DOI: 10.48550/arXiv.2301.04655.

7. Phuong, M. and M. Hutter \emph{Formal Algorithms for Transformers}.
2022. arXiv:2207.09238 DOI: 10.48550/arXiv.2207.09238.

8. Services, B.P., \emph{Introducing BloombergGPT, Bloomberg's
50-billion parameter large language model, purpose-built from scratch
for finance}. 2023.

9. Wu, S., et al. \emph{BloombergGPT: A Large Language Model for
Finance}. 2023. arXiv:2303.17564 DOI: 10.48550/arXiv.2303.17564.

10. Yang, X., et al., \emph{A large language model for electronic health
records.} NPJ Digit Med, 2022. \textbf{5}(1): p. 194.

11. Chung, H.W., et al. \emph{Scaling Instruction-Finetuned Language
Models}. 2022. arXiv:2210.11416 DOI: 10.48550/arXiv.2210.11416.

12. Ouyang, L., et al., \emph{Training language models to follow
instructions with human feedback.} ArXiv, 202\textbf{2.} abs/2203.02155.

13. Heaven, W.D. \emph{Why Meta's latest large language model survived
only three days online}. 2022; Available from:
\url{https://www.technologyreview.com/2022/11/18/1063487/meta-large-language-model-ai-only-survived-three-days-gpt-3-science/}.

14. Taylor, R., et al. \emph{Galactica: A Large Language Model for
Science}. 2022. arXiv:2211.09085 DOI: 10.48550/arXiv.2211.09085.

15. Chowdhery, A., et al. \emph{PaLM: Scaling Language Modeling with
Pathways}. 2022. arXiv:2204.02311 DOI: 10.48550/arXiv.2204.02311.

16. Zhang, S., et al. \emph{OPT: Open Pre-trained Transformer Language
Models}. 2022. arXiv:2205.01068 DOI: 10.48550/arXiv.2205.01068.

17. Touvron, H., et al. \emph{LLaMA: Open and Efficient Foundation
Language Models}. 2023. arXiv:2302.13971 DOI: 10.48550/arXiv.2302.13971.

18. \emph{Common Crawl}. Available from: \url{https://commoncrawl.org/}.

19. Mikolov, T., et al. \emph{Efficient Estimation of Word
Representations in Vector Space}. in \emph{International Conference on
Learning Representations}. 2013.

20. Pennington, J., R. Socher, and C. Manning. \emph{GloVe: Global
Vectors for Word Representation}. 2014. Doha, Qatar: Association for
Computational Linguistics.

21. Reimers, N. and I. Gurevych. \emph{Sentence-BERT: Sentence
Embeddings using Siamese BERT-Networks}. 2019. Hong Kong, China:
Association for Computational Linguistics.

22. Gururangan, S., et al. \emph{Don't Stop Pretraining: Adapt Language
Models to Domains and Tasks}. 2020. Online: Association for
Computational Linguistics.

23. Bender, E.M., et al., \emph{On the Dangers of Stochastic Parrots:
Can Language Models Be Too Big?}, in \emph{Proceedings of the 2021 ACM
Conference on Fairness, Accountability, and Transparency}. 2021,
Association for Computing Machinery: Virtual Event, Canada. p. 610--623.

24. Guu, K., et al., \emph{Retrieval Augmented Language Model
Pre-Training}, in \emph{Proceedings of the 37th International Conference
on Machine Learning}, D. Hal, III and S. Aarti, Editors. 2020, PMLR:
Proceedings of Machine Learning Research. p. 3929-\/-3938.

25. Lee, J., et al., \emph{BioBERT: a pre-trained biomedical language
representation model for biomedical text mining.} Bioinformatics, 2019.
\textbf{36}(4): p. 1234-1240.

26. Beltagy, I., K. Lo, and A. Cohan. \emph{SciBERT: A Pretrained
Language Model for Scientific Text}. 2019. Hong Kong, China: Association
for Computational Linguistics.

27. Gu, Y., et al., \emph{Domain-Specific Language Model Pretraining for
Biomedical Natural Language Processing.} ACM Trans. Comput. Healthcare,
2021. \textbf{3}(1): p. Article 2.

28. Chalkidis, I., et al. \emph{LEGAL-BERT: The Muppets straight out of
Law School}. 2020. Online: Association for Computational Linguistics.

29. Kaplan, J., et al. \emph{Scaling Laws for Neural Language Models}.
2020. arXiv:2001.08361 DOI: 10.48550/arXiv.2001.08361.

30. Ram, O., et al. \emph{In-Context Retrieval-Augmented Language
Models}. 2023. arXiv:2302.00083 DOI: 10.48550/arXiv.2302.00083.

31. Robertson, S. and H. Zaragoza, \emph{The Probabilistic Relevance
Framework: BM25 and Beyond.} Found. Trends Inf. Retr., 2009.
\textbf{3}(4): p. 333--389.

32. Ram, O., et al. \emph{Learning to Retrieve Passages without
Supervision}. 2022. Seattle, United States: Association for
Computational Linguistics.

33. Neelakantan, A., et al. \emph{Text and Code Embeddings by
Contrastive Pre-Training}. 2022. arXiv:2201.10005 DOI:
10.48550/arXiv.2201.10005.

34. OpenAI. \emph{ChatGPT}. Available from:
\url{https://chat.openai.com/}.

35. Microsoft, \emph{Bing.}

36. OpenAI \emph{GPT-4 Technical Report}. 2023. arXiv:2303.08774 DOI:
10.48550/arXiv.2303.08774.

37. NCBI. \emph{PMC}. Available from:
\url{https://www.ncbi.nlm.nih.gov/pmc/}.

38. NCBI, \emph{Entrez E-utilities.}

39. Dickinson, M.J., et al., \emph{Glofitamab for Relapsed or Refractory
Diffuse Large B-Cell Lymphoma.} New England Journal of Medicine, 2022.
\textbf{387}(24): p. 2220-2231.

40. Frieder, S., et al. \emph{Mathematical Capabilities of ChatGPT}.
2023. arXiv:2301.13867 DOI: 10.48550/arXiv.2301.13867.

41. Research, F. \emph{llama github page}. Available from:
\url{https://github.com/facebookresearch/llama}.

42. Mike Conover, M.H., Ankit Mathur, Xiangrui Meng, Jianwei Xie, Jun
Wan, Sam Shah, Ali Ghodsi, Patrick Wendell, Matei Zaharia and Reynold
Xin, \emph{Free Dolly: Introducing the World\textquotesingle s First
Truly Open Instruction-Tuned LLM}. 2023, Databricks.

43. \emph{Open Assistant}. Available from:
\url{https://open-assistant.io/}.

44. \emph{RedPajama, a project to create leading open-source models,
starts by reproducing LLaMA training dataset of over 1.2 trillion
tokens}. 2023.

45. Kojima, T., et al. \emph{Large Language Models are Zero-Shot
Reasoners}. 2022. arXiv:2205.11916 DOI: 10.48550/arXiv.2205.11916.

46. Wei, J., et al. \emph{Chain-of-Thought Prompting Elicits Reasoning
in Large Language Models}. 2022. arXiv:2201.11903 DOI:
10.48550/arXiv.2201.11903.

47. Press, O., et al. \emph{Measuring and Narrowing the Compositionality
Gap in Language Models}. 2022. arXiv:2210.03350 DOI:
10.48550/arXiv.2210.03350.

48. Yao, S., et al. \emph{ReAct: Synergizing Reasoning and Acting in
Language Models}. 2022. arXiv:2210.03629 DOI: 10.48550/arXiv.2210.03629.

49. Schick, T., et al. \emph{Toolformer: Language Models Can Teach
Themselves to Use Tools}. 2023. arXiv:2302.04761 DOI:
10.48550/arXiv.2302.04761.

50. Wu, C., et al. \emph{Visual ChatGPT: Talking, Drawing and Editing
with Visual Foundation Models}. 2023. arXiv:2303.04671 DOI:
10.48550/arXiv.2303.04671.

51. \emph{LangChain}. Available from:
\url{https://python.langchain.com/en/latest/}.

52. \emph{ChatGPT plugins}. Available from:
\url{https://openai.com/blog/chatgpt-plugins}.

53. Biderman, S., et al. \emph{Pythia: A Suite for Analyzing Large
Language Models Across Training and Scaling}. 2023. arXiv:2304.01373
DOI: 10.48550/arXiv.2304.01373.

54. Köpf, A., et al. \emph{OpenAssistant Conversations -\/-
Democratizing Large Language Model Alignment}. 2023. arXiv:2304.07327
DOI: 10.48550/arXiv.2304.07327.

55. Hu, E.J., et al. \emph{LoRA: Low-Rank Adaptation of Large Language
Models}. 2021. arXiv:2106.09685 DOI: 10.48550/arXiv.2106.09685.

56. Lester, B., R. Al-Rfou, and N. Constant \emph{The Power of Scale for
Parameter-Efficient Prompt Tuning}. 2021. arXiv:2104.08691 DOI:
10.48550/arXiv.2104.08691.

57. Li, X.L. and P. Liang. \emph{Prefix-Tuning: Optimizing Continuous
Prompts for Generation}. 2021. Online: Association for Computational
Linguistics.

58. Liu, X., et al. \emph{P-Tuning v2: Prompt Tuning Can Be Comparable
to Fine-tuning Universally Across Scales and Tasks}. 2021.
arXiv:2110.07602 DOI: 10.48550/arXiv.2110.07602.

59. Moor, M., et al., \emph{Foundation models for generalist medical
artificial intelligence.} Nature, 2023. \textbf{616}(7956): p. 259-265.

60. Acosta, J.N., et al., \emph{Multimodal biomedical AI.} Nat Med,
2022. \textbf{28}(9): p. 1773-1784.

\end{document}